\documentclass[lettersize,journal]{IEEEtran}
\usepackage{amsmath,amsfonts}
\usepackage{algorithmic}
\usepackage{algorithm}
\usepackage{array}
\usepackage[caption=false,font=normalsize,labelfont=sf,textfont=sf]{subfig}
\usepackage{textcomp}
\usepackage{stfloats}
\usepackage{url}
\usepackage{verbatim}
\usepackage{graphicx}
\usepackage{cite}
\newcommand{\tabincell}[2]{\begin{tabular}{@{}#1@{}}#2\end{tabular}} 
\usepackage{color}
\usepackage{multirow}
\usepackage[numbers,sort&compress]{natbib}
\usepackage{bbding}
\usepackage{booktabs}
\begin{document}

\title{Sparse-Dense Mixture of Experts Adapter for Multi-Modal Tracking}
\author{Yabin Zhu, Jianqi Li, Chenglong Li, \emph{Senior Member, IEEE}, Jiaxiang Wang, Chengjie Gu, Jin Tang

\thanks{
$\bullet$ Yabin Zhu, Jianqi Li and Chengjie Gu are with School of Public Security and Emergency Management, Anhui University of Science and Technology, Hefei 231131, China. (email: \{zhuyabin0726, lijianqi4129\}@foxmail.com, cjgu@aust.edu.cn)

$\bullet$ Chenglong Li is with Information Materials and Intelligent Sensing Laboratory of Anhui Province, Anhui Provincial Key Laboratory of Multimodal Cognitive Computation, School of Artificial Intelligence, Anhui University, Hefei 230601, China. (email: lcl1314@foxmail.com) 

$\bullet$ Jiaxiang Wang is with School of Artificial Intelligence, Anhui University of Science and Technology, Hefei 231131, China. (email: Netizenwjx@foxmail.com) 

$\bullet$ Jin Tang is with Information Materials and Intelligent Sensing Laboratory of Anhui Province, Anhui Provincial Key Laboratory of Multimodal Cognitive Computation, School of Computer Science and Technology, Anhui University, Hefei 230601, China. (email: tangjin@ahu.edu.cn)

$\bullet$ Corresponding author: Chenglong Li
}

\thanks{Manuscript received April 19, 2021; revised August 16, 2021.}}

\markboth{Journal of \LaTeX\ Class Files,~Vol.~14, No.~8, August~2021}%
{Shell \MakeLowercase{\textit{et al.}}: A Sample Article Using IEEEtran.cls for IEEE Journals}


\maketitle

\begin{abstract}
Parameter-efficient fine-tuning techniques, such as prompt and adapter, are widely used in multi-modal tracking due to their ability to solve some issues associated with full-model fine-tuning, such as time inefficiency, high resource demands, parameter storage burden and the risk of catastrophic forgetting. However, due to the cross-modal heterogeneity, most existing parameter-efficient fine-tuning multi-modal tracking methods struggle to represent multi-modal features well in a unified model framework and shared parameters.
To address this problem, we propose a novel Sparse-Dense Mixture of Experts Adapter (SDMoEA) fine-tuning framework that enables parameter-efficient multi-modal tracking tasks in a unified model framework and shared parameters.
In particular, we design a SDMoE module as the multi-modal adapter to effectively model modality-specific and shared information between modalities without demanding significant resources.
SDMoE consists of a sparse MoE and a dense-shared MoE, where the former primarily models the specific information in each fusion modality and the latter focuses on modeling shared information across modalities.
In addition, to address the shortcomings of existing tracking methods in modeling high-order correlations in multi-level, multi-modal feature fusion, we propose a Gram-based Semantic Alignment Hypergraph Fusion (GSAHF) module. This module first uses Gram matrices for cross-modal semantic alignment, ensuring that the constructed hypergraph accurately reflects the semantic similarity and high-order dependencies between modal features. It then incorporates the aligned features into the hypergraph structure, fully leveraging its inherent advantage in modeling high-order relationships to achieve deep fusion of multi-level, multi-modal information.
Extensive experiments demonstrate that our proposed method achieves superior performance compared to other parameter-efficient fine-tuning methods on various multi-modal tracking datasets, such as LasHeR, RGBT234, VTUAV, VisEvent, COESOT, DepthTrack and VOT-RGBD2022. 
\end{abstract}

\begin{IEEEkeywords}
Multi-Modal Tracking, Spare-Dense Mixture of Experts Adapter, Hypergraph Fusion.
\end{IEEEkeywords}

\section{Introduction}
\label{sec:intro}
\IEEEPARstart{V}{isual} tracking is a fundamental computer vision task that is widely used in scenarios such as autonomous driving, visual surveillance. Although RGB-based tracking has made significant progress, it faces challenges in complex scenarios such as low illumination, fast motion, and occlusion.
To address these challenges, a number of dual-modal tracking methods have been proposed, such as RGB-Thermal (RGB-T) tracking~\cite{lu2024breaking}, RGB-Event (RGB-E) tracking~\cite{wang2021viseventbenchmark}, and RGB-Depth (RGB-D) tracking~\cite{10623547}. 
However, these specialized methods are typically designed for a specific pair of modalities, lacking the capability to process diverse modalities within a unified framework.

\begin{figure}[tbp]
\centering 
\includegraphics[width=0.5\textwidth]{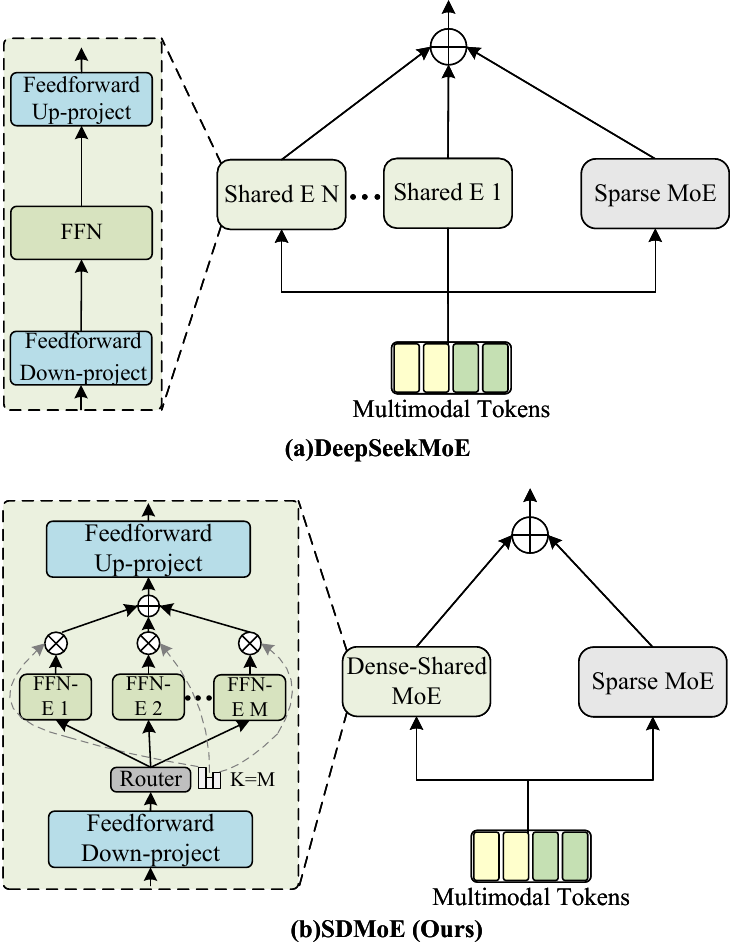}
   \caption{Comparison with different mixture of experts structures.}
   \label{fig:compare}
\end{figure}

To mitigate these problems, several parameter-efficient fine-tuning approaches~\cite{yang2022prompting, zhu2023visual, cao2024bi, hong2024onetracker, wu2024single, hou2024sdstrack} have been proposed for multi-modal tracking. These methods~\cite{yang2022prompting, zhu2023visual, cao2024bi, hong2024onetracker, hou2024sdstrack} adapt the pre-trained RGB model to multi-modal data using adapter or prompt techniques. While they achieve a unified framework, they fail to share the model parameters. As a result, separate models still need to be trained for each modality pair, which significantly compromises training efficiency. Liu et al.~\cite{10747517} propose an efficient multi-modal tracking framework that unifies network architecture and parameters. However, this approach requires training different modules of the network in multiple stages. It also necessitates fine-tuning the backbone network, which contradicts the design philosophy of parameter-efficient fine-tuning.
To alleviate the aforementioned issues, some works~\cite{wu2024single, chen2023unified} propose unified end-to-end training methods. However, the significant differences between modalities make these methods difficult to efficiently use the shared modal parameters to process multi-modal data, thereby limiting performance. Inspired by the successful application of Mixed of Experts (MoE) models in large language models, Tan et al.~\cite{tan2024xtrack} try to incorporate MoE networks into unified architectures for processing multi-modal data. Leveraging the strengths of MoE models, these approaches effectively model modality-specific information within a unified network architecture and parameter set, significantly improving model performance. However, these methods face the challenges of insufficient mining of modal shared information and inadequate modeling of higher-order information.
Recently, some full fine-tuning multi-modal tracking methods~\cite{chen2025sutrack, tan2025you} have also emerged, seeking to elevate multi-modal tracking performance to unprecedented heights. 
Nevertheless, such methods also introduce practical challenges, such as time inefficiency, high resource demands, parameter storage burden, and the risk of catastrophic forgetting.

To address the above problems, we propose a novel sparse-dense mixture of experts adapter fine-tuning framework that enables parameter-efficient multi-modal tracking in a unified end-to-end network. In particular, we design a SDMoE module, which is embedded as multi-modal adapter into a frozen RGB-based tracking network to effectively model modality-specific and shared information between different fused modalities without requiring significant resources. 
Specifically, the SDMoE consists of a sparse MoE and a dense-shared MoE. The sparse MoE contains $N$ experts, each responsible for modeling modality-specific information. The sparse MoE significantly expands model capacity without a substantial increase in computational effort, as only $K$ ($K < N$) specific experts are selected for each token during both training and testing.

The dense-shared MoE models share information between modalities. Unlike previous mixture of experts approaches~\cite{liu2024deepseek}, which use a parallel-serial structure, our design adopts a serial-parallel structure (as shown in Fig.~\ref{fig:compare}). The previous method~\cite{liu2024deepseek} faces a fundamental trade-off: employing multiple shared experts leads to prohibitive computational and memory costs, whereas using too few limits the model's ability to fully leverage inter-modal shared features. Our serial-parallel structure resolves this dilemma, significantly reducing resource consumption while ensuring effective utilization of cross-modal shared features. 
More specifically, all experts first process input features through a shared serial down-projection layer. The resulting representations are then processed in parallel by $M$ specific sub-networks (with $M=4$ in our implementation), each consisting of a single fully connected layer. Finally, outputs from these sub-networks are passed through a shared serial up-projection layer. This design ensures computational efficiency, as the parallel sub-networks operate only on the down-projected features, thereby keeping the overall computational load manageable.

Furthermore, to overcome the limitations of existing methods in modeling higher-order correlations during  multi-level, multi-modal feature fusion, this paper proposes a Gram-based semantic alignment hypergraph fusion module. This module first employs the Gram matrix to achieve cross-modal semantic alignment of multi-level, multi-modal features, ensuring that the subsequently constructed hypergraph accurately captures cross-modal semantic similarities and higher-order dependencies. Subsequently, the aligned features are organized into a hypergraph based on distance metrics. Hypergraph convolution is then applied to further extract higher-order relationships of multi-level, multi-modal features, thereby achieving more robust feature representations.

In summary, the contributions of this work are as follows:
\begin{itemize}
\item We propose a novel spare-dense mixture of experts adapter fine-tuning framework that enables parameter-efficient multi-modal tracking in a unified end-to-end trained network.
\item We design a SDMoE module, which mainly consists of a sparse MoE and a dense-shared MoE. The SDMoE can effectively model the specific and shared information between different fused modalities without requiring significant computational effort.
\item We also propose a Gram-based semantic alignment hypergraph fusion module to address the modeling of higher-order correlations in multi-level and multi-modal feature fusion, achieving more robust feature representations. 
\item Extensive experiments demonstrate that our proposed method achieves highly competitive performance against other parameter-efficient fine-tuning tracking methods across multiple multi-modal tracking datasets, including LasHeR, RGBT234, VTUAV, VisEvent, COESOT, DepthTrack, and VOT-RGBD2022.
\end{itemize}

\section{Related Work}
\subsection{Multi-modal Tracking}
In recent years, significant progress has been made in the field of multi-modal tracking, with numerous full fine-tuning tracking methods proposed~\cite{lu2024breaking, wang2021viseventbenchmark, 10623547}, leading to substantial performance improvements. In particular, certain unified multi-modal tracking approaches (such as SUTrack~\cite{chen2025sutrack} and FlexTrack~\cite{tan2025you} ) have elevated visual tracking to new heights.
However, these algorithms often suffer from time inefficiency, high resource demands, parameter storage burdens, and the risk of catastrophic forgetting. To address these challenges, several parameter-efficient fine-tuning tracking methods~\cite{yang2022prompting, zhu2023visual, hong2024onetracker, cao2024bi, hou2024sdstrack} have emerged. ProTrack~\cite{yang2022prompting} is the first to introduce the concept of prompt learning into multi-modal tracking, but it lacks effective fine-tuning. Furthermore, ViPT~\cite{zhu2023visual} employs a prompt fine-tuning method in multi-modal tracking and achieves competitive tracking performance at that time. SDSTrack~\cite{hou2024sdstrack} proposes a complementary masked patch distillation strategy to improve tracking robustness. However, these trackers require training a separate model for each pair of modalities, leading to lower training efficiency.

Uni-Track~\cite{wu2024single} and OneTracker~\cite{hong2024onetracker} attempt to alleviate this problem by using a unified tracking framework and parameters. However, the significant differences between modalities make it difficult for shared modal parameters to efficiently process multi-modal data, therefore limiting tracking performance. Inspired by the successful application of Mixed of Experts (MoE) models in large language models, XTrack~\cite{tan2024xtrack} attempts to incorporate MoE networks into the RGB framework for modeling modality-specific information within unified network architectures and parameter sets. However, these methods also face challenges in extracting shared information between modalities and inadequately modeling higher-order information.
\begin{figure*}[htbp]
\centering 
\includegraphics[width=1\textwidth]{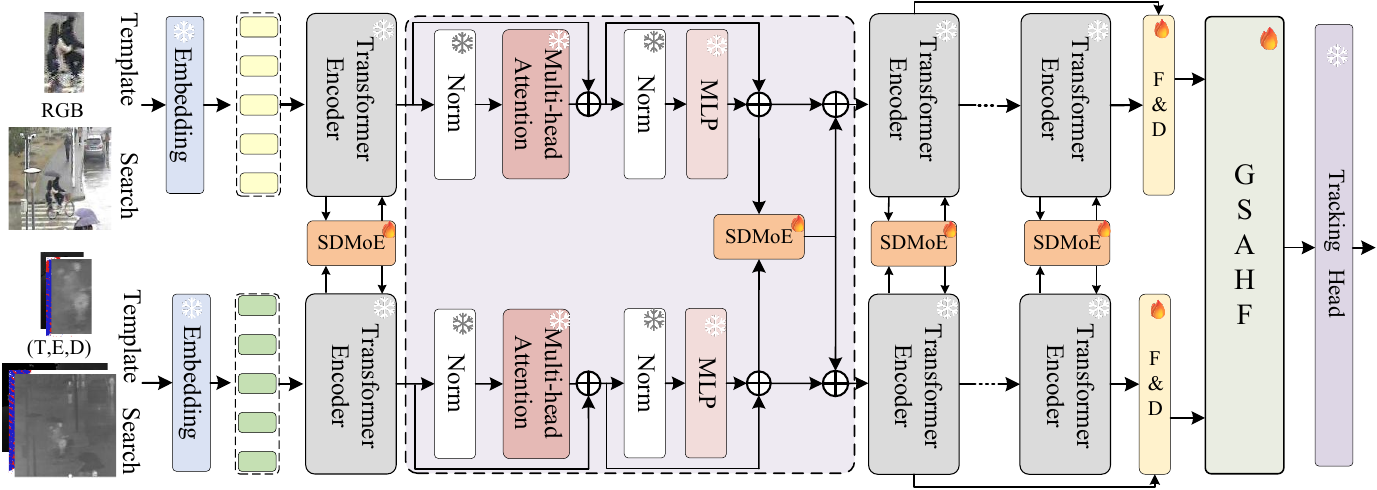}
   \caption{Overview architecture of our proposed method. Here, we embed the Spare-Dense Mixture of Experts (SDMoE) module as the multi-modal adapter to effectively model modality-specific and shared information of fused modalities. 
   Moreover, we also embed the Gram-based Semantic Alignment Hypergraph Fusion (GSAHF) module to address the modeling of higher-order correlations in multi-level and multi-modal feature fusion, achieving more robust feature representations.}
   \label{fig:framework}
\end{figure*}
\subsection{Mixture of Experts}
Jacobs et al.~\cite{jacobs1991adaptive} first propose the mixture of experts, which serves as a model architecture that effectively improves the adaptability of models when dealing with diverse tasks by dividing the work among different expert modules. Sparse MoE~\cite{lepikhin2020gshard} is an improved version that activates only a small number of expert modules to reduce computational overhead, and has been widely used in Large Language Models (LLMs), particularly performing well in large-scale training environments.
In recent years, on the basis of sparse MoE, Dai et al.~\cite{dai2024deepseekmoe, liu2024deepseek} introduce shared experts, in which shared experts are specifically used to capture common knowledge across tasks, thereby reducing redundancy among experts. This approach improves parameter efficiency and reduces redundancy between experts. Inspired by the above methods, some works~\cite{cai2025spmtrack, tan2024xtrack} introduce MoE module into visual tracking, achieving significant performance improvements.
However, due to the significant differences between multi-modal data, directly using sparse MoE with shared experts faces challenges: it either requires multiple parallel shared experts, leading to excessive computational resource consumption and an excessive number of parameters, or the use of a single shared expert, which makes it difficult to fully utilize the commonality of information between different modalities.

\subsection{Hypergraph Learning}
A hypergraph~\cite{zhou2006learning} consists of a set of vertices and hyperedges. Unlike ordinary graphs, each edge in a hypergraph can connect multiple vertices, enabling the modeling of more complex, higher-order data relationships. Hypergraphs have been widely applied in fields such as epidemic spreading~\cite{antelmi2020design} and social networks~\cite{xiao2024information}. HGNN~\cite{feng2019hypergraph} leverages this structural relationship to enable direct learning from hypergraphs through spectral convolutions. The subsequent HGNN+~\cite{gao2022hgnn+} further introduces higher-order information propagation mechanisms between vertices, expanding hypergraph learning capabilities from a spatial perspective. In computer vision, HyperYOLO~\cite{10818703} first introduces hypergraph convolutions to object detection tasks for modeling higher-order relationships between features, significantly enhancing object detection performance. 
Building upon this, YOLOv13~\cite{lei2025yolov13} optimises hypergraph construction through an adaptive hyperedge generation module, enabling more precise higher-order feature modeling and further improving detection accuracy. While these methods show strong performance in single-modal visual tasks, direct hypergraph computation struggles to effectively model higher-order relationships between features across modalities in multi-modal tracking tasks due to significant inter-modal differences.

\section{Methodology}
\label{sec:methodology}
In this section, we first introduce the overall architecture of our SDMoEA. Next, we provide detailed information on the SDMoE and the GSAHF module. Finally, we present the loss function details of our method.

\subsection{Overview}
The overall framework of the proposed method is illustrated in Fig.~\ref{fig:framework}. The proposed multi-modal tracking framework is an extension of the RGB tracking network~\cite{zheng2024odtrack}, which adopts a typical dual-stream structure to introduce an additional modality X (such as Thermal, Depth, or Event) alongside the RGB modality. Given the search regions and template images from two modalities, they are first embedded into patches by a patch embedding layer and flattened into one-dimension tokens. It should be noted that our method is based on ODTrack~\cite{zheng2024odtrack}. Therefore, we retain its input design that introduces multi-frame images as input to leverage temporal information. Next, these tokens are fed into the ViT~\cite{dosovitskiy2020image}, which employs the token elimination mechanism proposed in OSTrack~\cite{ye2022joint} for feature extraction. 
During the feature extraction process, as shown in the middle part of Fig.~\ref{fig:framework}, we integrate the proposed SDMoE module into each Transformer block to enhance the model's ability to adapt to multi-modal input data. The specific design and meaning of the SDMoEA will be elaborated in the next chapter. 
Next, we design a Gram-based semantic alignment hypergraph fusion module, which can model higher-order correlations of multi-modal, multi-level features for more robust feature representation. 
Finally, the enhanced features are fed into the box head layer for predicting the bounding box of the target. 

\begin{figure*}[htbp]
\centering 
\includegraphics[width=1\textwidth]{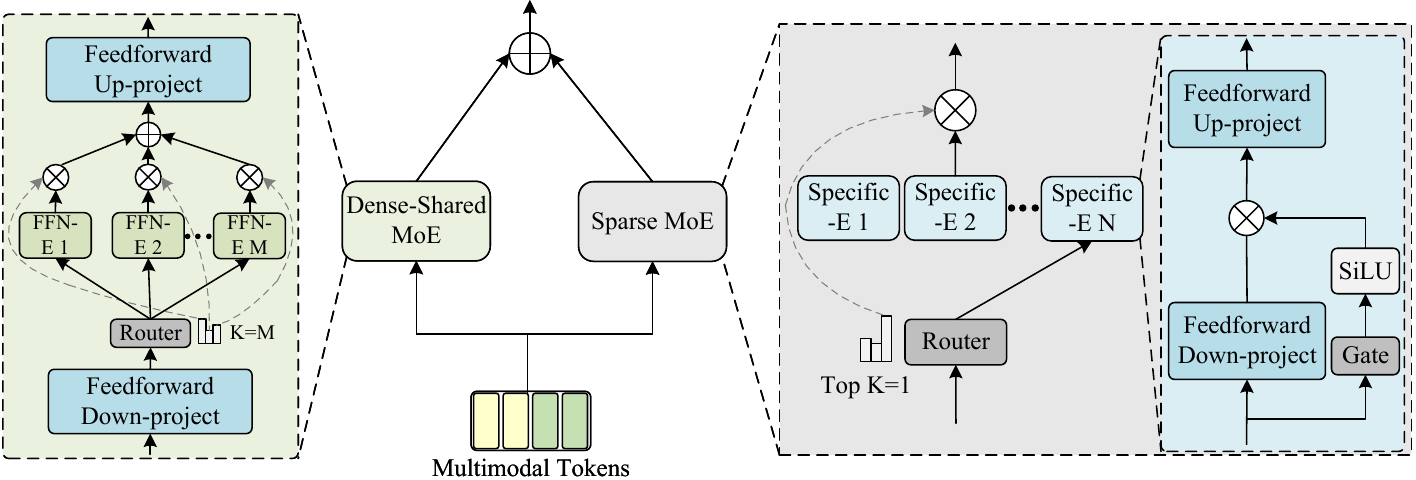}
   \caption{Detailed design of the proposed SDMoE module. The SDMoE consists of a spare MoE and a dense-shared MoE. Spare MoE has a router and $N$ specific experts. The details of specific expert is shown on the right of the figure. Correspondingly, details about dense-shared MoE are shown on the left of the figure. Specific-E and FFN-E denote specific expert and the feedforward expert network of dense-shared MoE, respectively. The SiLU is a activation function, called Sigmoid Linear Unit.}
   \label{fig:moe}
\end{figure*}
\subsection{Sparse-Dense Mixture of Experts}
Recently proposed parameter-efficient fine-tuning tracking methods ~\cite{cao2024bi, hong2024onetracker, hou2024sdstrack, wu2024single, tan2024xtrack} address the key shortcomings of full fine-tuning tracking methods, including time inefficiency, high resource consumption, significant parameter storage burden, and catastrophic forgetting, and have achieved notable success. However, constrained by the significant differences between modalities, these approaches remain challenged within the framework of shared parameters to effectively model both the specific features of each modality and the cross-modal shared features simultaneously.

To address the problem, we propose a SDMoE module as multi-modal adapter, which can efficiently integrate multi-modal data in an end-to-end trained network. Specifically, as shown in the middle part of Fig.\ref{fig:framework}, the SDMoE is embedded after the MLP layer of each ViT block, effectively modeling the shared and specific features of different fused modalities. As shown in Fig.~\ref{fig:moe}, the SDMoE mainly consists of a spare MoE and a lightweight dense-shared MoE.

\subsubsection{Sparse MoE} 
The sparse MoE contains a router and $N$ specific experts, which is mainly used to model specific information in modalities. The router consists of a fully connected layer and a $Softmax$ function. The specific operation can be formalized as,
\begin{equation}
\begin{aligned}
s_{t,n} = Softmax(FFN(Token_{t},N))
\end{aligned}
\end{equation}
where $s_{t,n}$ denotes the score value of the t\emph{-th} token corresponding to the n\emph{-th} expert. The $N$ is the total number of specific experts. In this paper, the value of $N$ is 4. The $FFN$ represents the fully connected layer. Auto-learned routing strategies may run the risk of routing collapse, that is, the model consistently selects only a small number of experts, preventing others from being adequately trained. To address this problem, following~\cite{dai2024deepseekmoe}, we use expert-level balance losses. The computation of the balance loss is as follows:
\begin{equation}
\begin{aligned}
{L}_{\text {eb}} & =\sum_{n=1}^{N} f_n P_n \\
f_n & =\frac{N}{KT} \sum_{t=1}^T \mathbf{1}(\text { Token } t \text { selects Expert } n), \\
P_n & =\frac{1}{T} \sum_{t=1}^T s_{t,n}
\end{aligned}
\end{equation}
where $T$ is the total number of tokens, and $K$ is the number of experts selected for each forward propagation of the network. In this paper, the value of $K$ is 1.
The $\mathbf{1}(\cdot)$ represents the indicator function, which is as a Boolean judgment condition, that is, if the token $t$ is assigned to the expert $n$, then the value of the indicator function is 1, otherwise the value is 0.

As shown in the right part of Fig.~\ref{fig:moe}, each specific expert consists of a feedforward down-project layer, feedforward up-project layer, gate layer, and activation function for each specific expert. This feedforward down-up project structure significantly reduces computational overhead. The feature dimensions input to the feedforward down-project layer here are $[B, H, D]$. After processing by this feedforward down-project layer, its features become $[B, H, D/G]$, and this $G$ is taken as 12. The network structures of gate layer and the feedforward down-project layer are consistent, and the activation function used is SiLU~\cite{elfwing2018sigmoid}. It should be emphasized that the sparse MoE significantly expands the model capacity, allowing it to model the specific information of different modalities well. However, its computational effort does not increase significantly, mainly because only one specific expert is selected to participate in the forward propagation during training and testing.

\subsubsection{Dense-Shared MoE} 
For the dense-shared MoE, it focuses on modeling the shared information in modalities. As shown in the left part of Fig.~\ref{fig:moe}, unlike previous multi-shared experts~\cite{liu2024deepseek} that use a parallel-serial structure, our proposed dense-shared MoE adopts a serial-parallel design. This design effectively solves the problems of excessive computational resource consumption caused by multiple parallel shared experts and overcomes the challenge of a single shared expert failing to fully exploit the shared features among modalities. Here, the parallel-serial structure of multiple shared experts refers to the fact that multiple experts are parallel and the network structure within each expert is serial; whereas our serial-parallel structure means that all experts go through the shared serial feedforward down-project layer and feedforward up-project layer, and the parallel part is the unique sub-network of each expert. 

Specifically, the multi-modal tokens are fed to feedforward down-project layer for features dimensionality reduction. Similarly to a specific expert, the input features will be reduced in dimensionality from $[B, H, D]$ to $[B, H, D/G]$. The dimensionally reduced features will be fed to the router and the parallel shared feedforward expert sub-network to extract the shared features of the different modalities. The extracted shared features will be weighted and fused with the expert weights generated by the routers and the final modal shared features will be obtained through the feedforward up-project layer. Here, there are $M$ parallel shared feedforward expert and $M$ is set to 4, and each parallel shared expert consists of only one fully connected layer. It is worth noting that this series-parallel structure will require significantly less computation by virtue of its low dimensional features and less parallel parameters.

\begin{figure}[htbp]
\centering 
\includegraphics[width=0.49\textwidth]{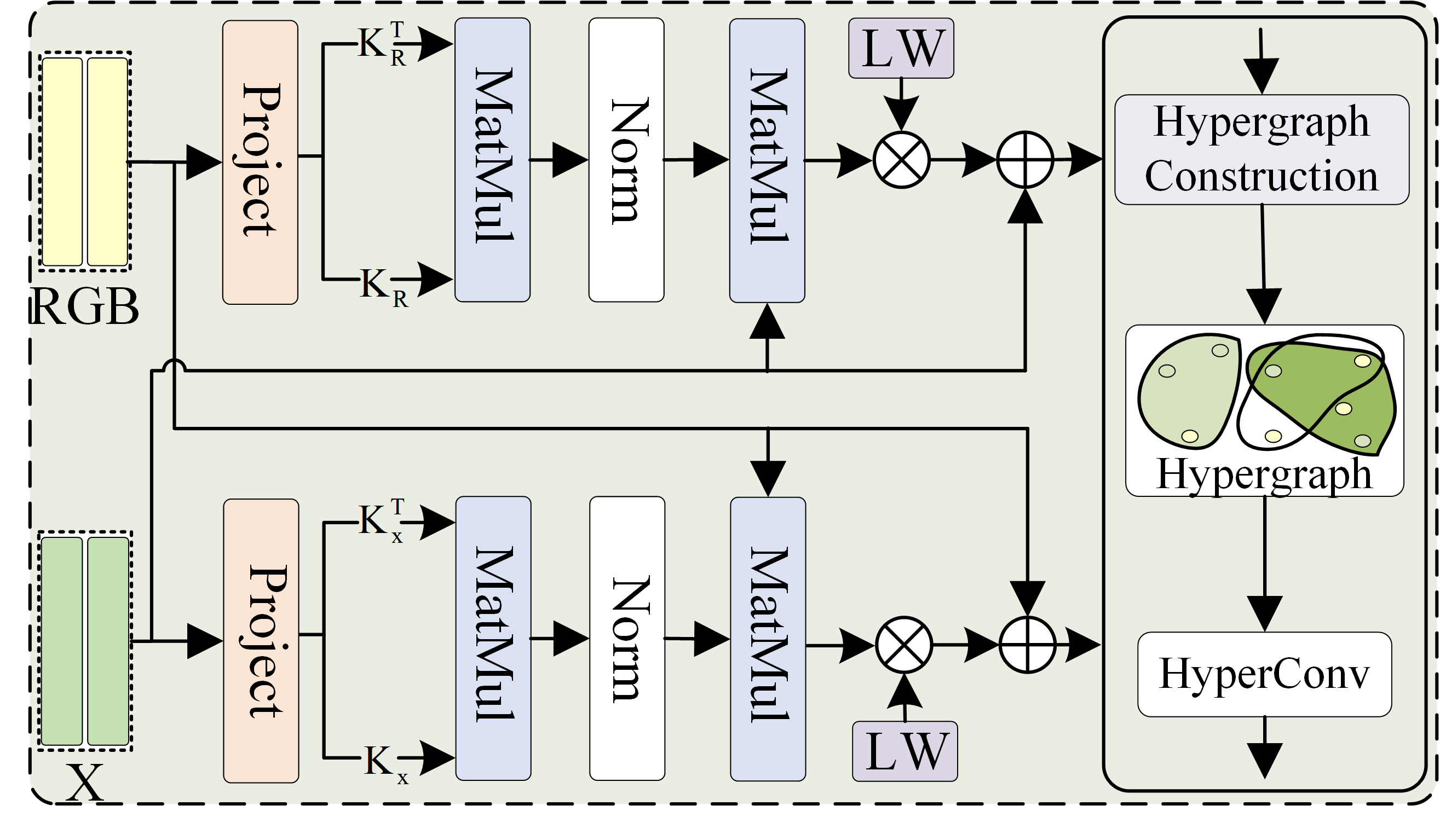}
   \caption{The detailed design of the proposed GSAHF module. The $\mathbf{LW}$ denotes learnable weights.}
   \label{fig:gsma}
\end{figure}

\begin{table*}[hbp]\small
  \centering
\renewcommand\arraystretch{1}  
 \caption{Overall performance on LasHeR and RGBT234. Results are reported in percentage (\%). PEFT and FFT represent parameter-efficient fine-tuning and full fine-tuning, respectively. The $*$ indicates that our method performs full fine-tuning.}
     \setlength{\tabcolsep}{5mm}{ \begin{tabular}{c|c|l|c|rr|rr}
    \toprule
    \multicolumn{2}{c|}{\multirow{2}[2]{*}{Method Type}} & \multicolumn{1}{c|}{\multirow{2}[2]{*}{Method}} & \multicolumn{1}{c|}{\multirow{2}[2]{*}{Source}} & \multicolumn{2}{c|}{LasHeR} & \multicolumn{2}{c}{RGBT234} \\
    \multicolumn{2}{c|}{} &      &      & \multicolumn{1}{c}{PR} & \multicolumn{1}{c|}{AUC} & \multicolumn{1}{c}{MPR} & \multicolumn{1}{c}{MSR} \\
    \midrule
    \multicolumn{1}{c|}{\multirow{19}[4]{*}{\tabincell{c}{Multi-modal \\ Trackers}}} & \multicolumn{1}{c|}{\multirow{11}[2]{*}{PEFT}} & SDMoEA-L &  \textbf{Ours} & \textbf{77.7} & 60.9 & \textbf{92.6} & 68.3 \\
         &      & SDMoEA-B & \textbf{Ours} & 76.5 & 60.3 & 92.2 & \textbf{68.7} \\
         &      & SeqTrackv2-L~\cite{chen2023unified}    &  arXiv'23    &   76.7   &   \textbf{61.0}   &   91.3   & 68.0 \\ 
         &      & SeqTrackv2-B~\cite{chen2023unified}   &  arXiv'23    &  71.5    &  56.2    &   90.0   &  66.3\\          
         &      & XTrack-L~\cite{tan2024xtrack}    &  ICCV'25    &  73.1    &  58.7    &  87.8    & 65.4 \\
         &      & XTrack-B~\cite{tan2024xtrack}    &  ICCV'25 &  69.1    &  55.7    &  87.4    & 64.9 \\       
         &      & OneTracker~\cite{hong2024onetracker} & CVPR'25     & 67.2 & 53.8 & 85.7 & 64.2 \\
         &      & EMTrack~\cite{10747517}   &  TCSVT'25   &    65.9   &   53.3   &   81.8   & 60.1 \\         
         &      &Un-Track~\cite{wu2024single}    &  CVPR'24    &  64.6    &   51.3   &   84.2   & 62.5 \\
         &      &SDSTrack~\cite{hou2024sdstrack}    &  CVPR'24    &  66.5    &   53.1  &   84.8   & 62.5 \\       
         &      &ViPT~\cite{zhu2023visual}    &  CVPR'23    &  65.1    &   52.5  &   83.5   & 61.7 \\
\cmidrule{2-8}         & \multirow{9}[2]{*}{FFT} & SDMoEA*-L & \textbf{Ours} &\textbf{79.5} & \textbf{62.9}&\textbf{93.8} & 69.0 \\
         &      & SDMoEA*-B & \textbf{Ours} & 77.3 & 61.0 & 93.5 &\textbf{70.3} \\
         &      & SUTrack-L~\cite{chen2025sutrack}  & AAAI'25& 76.9 & 61.9& 93.7 & \textbf{70.3} \\
         &      & SUTrack-B~\cite{chen2025sutrack} & AAAI'25 &75.8 & 60.9 & 92.1 & 69.2 \\
         &      & FlexTrack~\cite{tan2025you} &   ICCV'25   & 77.3 & 62.0 & 92.7 & 69.9 \\
         &      & STTrack~\cite{hu2025exploiting} &    AAAI'25     &76.0 & 60.3 & 89.8 &66.7 \\
         &      &  APTrack~\cite{aptrack}    & TAI'25    &  74.1    &   58.9   &   90.6   & 67.2 \\
         &      & CMDTrack~\cite{cmdtrack}     & PAMI'25     & 68.8     & 56.6     & 85.9     & 61.8 \\
    \midrule
    \multirow{14}[3]{*}{\tabincell{c}{RGBT \\ Trackers}} & \multicolumn{1}{c|}{\multirow{7}[2]{*}{PEFT}} & LRPD-L~\cite{hu2025exploiting2}    &  ICMR'25    &  \textbf{80.5}    &  \textbf{63.8}    &  \textbf{93.6}    &\textbf{70.8} \\
         &      & LRPD-B~\cite{hu2025exploiting2}    &  ICMR'25 &  78.3    &  62.0    &   92.6   &  69.5      \\
         &      & MPANet~\cite{liu2025scale}   &  TMM'25    &  72.1    &   57.3   &  87.4    & 64.6 \\
         &      & TATrack~\cite{wang2024temporal}   &  AAAI'24    &  72.1    &   56.1   &  87.2    & 64.4 \\         
         &      & BAT~\cite{cao2024bi}   &  AAAI'24    &  70.2    &   56.3   &  86.8    & 64.1 \\
         &      &TIPTrack~\cite{yan2025tiptrack}   &ESWA'25    &  71.2    &   56.1   &  89.2    & 65.3 \\         
         &      & MAT~\cite{wang2025multi}    &  INFFUS'25    & 67.0     & 53.7     &   84.5   &62.7  \\

\cmidrule{2-8}         & \multirow{6}[1]{*}{FFT}    & CKD~\cite{lu2024breaking}     &ACM MM'24      &  \textbf{73.2}     & 58.1     &  90.0 & \textbf{67.4}    \\
         &      & AINet~\cite{lu2025rgbt}    & AAAI'25      & 73.0     &  \textbf{58.2}     &  89.1    & 66.8 \\
         &      & CAFormer~\cite{xiao2025cross}    &  AAAI'25    &  70.0    & 55.6     &  88.3    & 66.4 \\
         &      & MOETrack~\cite{tang2025revisiting}    &  TIP'25    &  72.1    & 57.8     &  88.1    & 65.1 \\         
         &      & IPLL~\cite{lu2025modality}    &  IJCV'25    & 69.4     & 55.3     &  88.3    & 65.7 \\
         &      & AFTER~\cite{11079869}    &   TIP'25   &  70.3    & 55.1     &  \textbf{90.1}     & 66.7 \\

\bottomrule
    \end{tabular}}
  \label{tab:rgbt}%
\end{table*}%

\subsection{Gram-based Semantic Alignment Hypergraph Fusion}
The high-order relationships between features are crucial for computer vision tasks~\cite{feng2019hypergraph, gao2022hgnn+} dealing with complex scenes, yet existing tracking methods often overlook this issue. Inspired by certain object detection approaches~\cite{feng2019hypergraph, gao2022hgnn+}, we consider employing hypergraphs to model the complex high-order relationships of multi-level, multi-modal features. However, practical experimentation reveals that directly applying hypergraph modeling to multi-modal data struggles to accurately capture intrinsic connections between different modal features due to inherent modal differences. To address the problem, we design a Gram-based semantic alignment hypergraph fusion module. 

Specifically, inspired by~\cite{moda}, this module employs Gram matrices to achieve cross-modal semantic alignment, ensuring that the constructed hypergraph accurately captures both semantic similarities and high-order dependencies among multi-level, multi-modal features. The multi-level, multi-modal features are constructed by first extracting features from shallow to deep layers of each modality and then integrating them across layers via fully connected layers to build a more robust feature representation. As shown in Fig.~\ref{fig:gsma}, the Gram matrix is first computed from the key vectors $K$ of tokens within each modality. Taking the RGB modality as an example, the Gram matrix is defined as follows:
\begin{equation}
G_{R}=K_{R}^{\top} K_{R}
\end{equation}
The $K_{R}$ represents the key vector of the RGB modality. The Gram matrix $G_{R}$ is subsequently normalized to form the basis vectors of the RGB modality space, effectively acting as a transition operator. Based on the Gram basis vectors, tokens $K_{X}$ from any other modality X can be effectively mapped to the RGB modality, thereby reducing the inter-modal semantic gap and achieving semantic alignment between the two modalities. The specific operations are as follows:
\begin{equation}
K_{X \rightarrow R}=K_{X}\left\|G_R\right\|
\end{equation}
The mapped modal tokens $K_{X \rightarrow R}$ are further fused with the original modal tokens to enhance token similarity between modalities. The procedure is as follows:
\begin{equation}
F_{X}=K_{X}+W_{X}*K_{X \rightarrow R}
\end{equation}
where $W_{X}$ is a learnable weight. Similarly, we perform an interactive operation between the two modalities. Taking the X modality as another example, the fused modality formed after transformation from the RGB modality to the X modality can be obtained using the following formula.
\begin{equation}
F_{R}=K_{R}+W_{R}*K_{R \rightarrow X}
\end{equation}
Finally, $F_{X}$ and $F_{R}$ are concatenated and fed into a fully connected layer to further enhance the similarity between modalities.

After obtaining the aligned multi-level multi-modal features, we employ a hypergraph structure to further enhance their high-order modeling capability. Specifically, a hypergraph \( G \) is constructed, consisting of a vertex set \( V \) and a hyperedge set \( E \). We use the grid-structured visual features as the vertex set \( V \) of the hypergraph and construct hyperedges based on a distance threshold in the feature space. The overall hyperedge set can be formally defined as:
\begin{equation}
\mathcal{E} = \{ \operatorname{ball}(v, \epsilon) \mid v \in V \}
\end{equation}
where $\operatorname{ball}(v, \epsilon) = \left\{ u \mid \| \boldsymbol{x}_u -\boldsymbol{x}_v \|_2 < \epsilon, u \in V \right\}$ 
denotes the set of neighboring vertices whose Euclidean distance to vertex \( v \) is less than the threshold \( \epsilon \), and \( \| \boldsymbol{x} - \boldsymbol{y} \|_2 \) is the Euclidean distance function. Given a vertex set \( V \) and a hyperedge set \( E \), an incidence matrix \( H \) can be generated based on the relationships between them. During computation, the hypergraph \( G \) is generally represented by its incidence matrix \( H \).
\begin{equation}
H(i, e)= \begin{cases}1 & \text { if } v_i \in e \\ 0 & \text { otherwise }\end{cases}
\end{equation}

To achieve higher-order message passing on hypergraph structures, we adopt a typical spatial domain hypergraph convolutional layer~\cite{gao2022hgnn+} and incorporate residual connections and  trainable parameter $\boldsymbol{\Theta_2}$  to enable higher-order learning of vertex features, as illustrated below:
\begin{equation}
\text { HyperConv }(\boldsymbol{X}, \boldsymbol{H})=\boldsymbol{X}+\boldsymbol{\Theta_2} \boldsymbol{D}_v^{-1} \boldsymbol{H} \boldsymbol{D}_e^{-1} \boldsymbol{H}^{\top} \boldsymbol{X} \boldsymbol{\Theta_1}
\end{equation}
where $\boldsymbol{D}_v$ and $\boldsymbol{D}_e$ represent the diagonal degree matrices of
the vertices and hyperedges, respectively. $\boldsymbol{X}$ is a vertex set feature, i.e., a multi-level, multi-modal feature. $\boldsymbol{\Theta_1}$ and $\boldsymbol{\Theta_2}$ are trainable parameters.

\subsection{Loss Function}
Following ODTrack, we adopt the weighted focal loss~\cite{law2018cornernet} for classification, $l_1$ loss and generalized IoU loss~\cite{rezatofighi2019generalized} for bounding box regression. Following~\cite{dai2024deepseekmoe}, we adopt expert-level balance losses to solve the problem of routing crashes. The total loss function can be formulated as follows:
\begin{equation}
\begin{aligned}
L_{\text {total}}=L_{cls}+\lambda_{\text {iou }} L_{i o u}+\lambda_{L_1} L_1 + \alpha  {L}_{\text {eb}}
\end{aligned}
\end{equation}
where $\lambda_{\text {iou }}$, $\lambda_{L_1}$, and $\alpha$ are hyper-parameters, and are set to 2, 5, and 0.001, respectively.

\section{Experiment}
In this section, we first describe the specific implementation details of the proposed method, followed by a detailed quantitative evaluations. The evaluations cover the performance of the proposed method and the comparison method on RGB-E, RGB-T, and RGB-D tracking datasets, as well as the results of a series of ablation studies. Finally, we also show some results of the qualitative evaluations through visualization in order to present the performance and effectiveness of the proposed method more intuitively.

\subsection{Implementation Details}
This section presents a detailed exposition of the implementation details for the proposed method, including the specific configurations for both the training and inference phases.
\subsubsection{Training}
Our proposed method is built upon the ODTrack architecture~\cite{zheng2024odtrack}, which includes two versions of feature extractors: one based on a ViT-B backbone with 12 Transformer layers, and the other based on a ViT-L backbone with 24 Transformer layers. We load the RGB-based pre-trained model of ODTrack, and then fine-tune it by using multi-modal datasets, including the training datasets of LasHeR~\cite{li2021lasher} and VTUAV~\cite{zhang2022visible}, VisEvent~\cite{wang2021viseventbenchmark}, COESOT~\cite{tang2022revisiting}, DepthTrack~\cite{yan2021depthtrack}, and RGBD1K~\cite{zhu2023rgbd1k}. Note that we only train a model to cope with the multi-modal tracking task.
Specifically, the proposed network is trained on two NVIDIA A800 GPUs, with batch sizes set to 8 for the ViT-B model and 4 for the ViT-L model, respectively. All experiments of our method are conducted using PyTorch-1.10.2+cu113. The optimization process employs the AdamW optimizer~\cite{ilya2019decoupled}, with a weight decay parameter set to $1\times10^{-4}$. We load pre-trained models of ODTrack and train our network with a learning rate of $4\times10^{-5}$ for a total of 35 epoch for ViT-B model and 40 epoch for the ViT-L model. It should be emphasized that we will freeze the foundation model and train only the newly added module.

\subsubsection{Inference}
After training completes, both model parameters and hyperparameters remain fixed and are employed for evaluation on multimodal tracking test datasets. All multimodal tests are conducted using the identical set of models and hyperparameters. During the inference stage, following the baseline approach, we provide the tracker with four frames including the initial frame as reference templates to fully leverage temporal information and enhance tracking robustness. Furthermore, our tracking model based on the ViT-B architecture, with 118.28 million parameters and a computational complexity of 189.47 GFLOPs, achieves a tracking speed of approximately 19.3 FPS on an NVIDIA A800 GPU.

\subsection{Quantitative Evaluations}
This comparative analysis primarily covers two major categories of methods: unified multi-modal tracking approaches and those specifically designed for RGB-X modality pairs. Within each category, we further subdivide into two strategies: Parameter-Efficient Fine-Tuning (PEFT) and Full Fine-Tuning (FFT). To demonstrate the potential of our proposed method, we also present a full fine-tuning version of our approach, named SDMOEA*.
\begin{table}[htbp]\small
  \centering
\renewcommand\arraystretch{1.2}  
     \caption{Overall performance on VisEvent. Results are reported in percentage (\%). PEFT and FFT represent parameter-efficient fine-tuning and full fine-tuning, respectively. The $*$ indicates that our method performs full fine-tuning.}
     \setlength{\tabcolsep}{1mm}{ \begin{tabular}{c|c|l|c|cc}
    \toprule
    \multicolumn{2}{c|}{\multirow{2}[2]{*}{Method Type}} & \multicolumn{1}{c|}{\multirow{2}[2]{*}{Method}} & \multicolumn{1}{c|}{\multirow{2}[2]{*}{Source}} & \multicolumn{2}{c}{VisEvent}  \\
    \multicolumn{2}{c|}{} &      &      & \multicolumn{1}{c}{PR} & \multicolumn{1}{c}{AUC} \\
    \midrule
    \multicolumn{1}{c|}{\multirow{20}[4]{*}{\tabincell{c}{Multi-modal \\ Trackers}}} & \multicolumn{1}{c|}{\multirow{11}[2]{*}{PEFT}} & SDMoEA-L &  \textbf{Ours} & \textbf{81.3} & 61.9 \\
         &      & SDMoEA-B & \textbf{Ours} & 80.3 & 61.2\\
         &      & SeqTrackv2-L    &  arXiv'23    &   80.0   &   \textbf{63.4}  \\ 
         &      & SeqTrackv2-B   &  arXiv'23    &  79.3    &  62.2  \\          
         &      & XTrack-L   &  ICCV'25   &80.5 &  63.3  \\
         &      & XTrack-B    &  ICCV'25 &77.5   &60.9   \\       
         &      & OneTracker & CVPR'25     &76.7 &60.8 \\
         &      & EMTrack   &  TCSVT'25   &72.4  &58.4\\         
         &      &Un-Track   &  CVPR'24    &  75.5    &   58.9 \\
         &      &SDSTrack   &  CVPR'24    &76.7    &59.7\\       
         &      &ViPT    &  CVPR'23    &75.8    &59.2 \\
\cmidrule{2-6}         & \multirow{9}[2]{*}{FFT} & SDMoEA*-L & \textbf{Ours} &\textbf{82.2} & 63.8 \\
         &      & SDMoEA*-B & \textbf{Ours} & 81.3 & 61.9 \\
         &      & SUTrack-L  & AAAI'25& 80.5 & 63.8  \\
         &      & SUTrack-B & AAAI'25 &79.8& 63.4\\
         &      & FlexTrack&   ICCV'25   &81.4 &\textbf{64.1}\\
         &      & STTrack &    AAAI'25     &78.6& 61.9\\
         &      & APTrack   & TAI'25    &  78.5    &61.8\\
         &      & CMDTrack    & PAMI'25     &75.8    &61.3  \\
    \midrule
    \multirow{5}[3]{*}{\tabincell{c}{RGBE \\ Trackers}} &\multicolumn{1}{c|}{\multirow{1}[2]{*}{PEFT}}   & EventTPT    &  ICRA'25   &  75.7   &  60.0   \\  
    \cmidrule{2-6}         & \multirow{4}[2]{*}{FFT}  & CRSOT   &  TMM'25    &  \textbf{74.1}    &  \textbf{52.5}    \\
    && CEUTrack   &  PR'25    &  69.1    &  53.1\\

    &  &Siam RCNN-FE   & CVPR'20   &64.0   & 49.9  \\     
    &  &VisEvent-MDNet   & TC'23   &63.2   & 43.0   \\     
\bottomrule
    \end{tabular}}
  \label{tab:rgbe}%
\end{table}%

\subsubsection{Evaluation on RGB-T Tracking Datasets}
In multi-modal tracking tasks, RGB-T tracking has garnered significant attention, with numerous comparative methods available for reference. To this end, we conduct comparative experiments on two widely used large-scale RGB-T tracking datasets, including RGBT234~\cite{li2019rgb} and LasHeR. As shown in Table~\ref{tab:rgbt}, our proposed method achieves state-of-the-art performance compared to existing PEFT approaches. Even when compared to other FFT methods, our PEFT version demonstrates strong competitiveness. Furthermore, our FFT version achieves the best performance among all FFT methods, attaining state-of-the-art results. In particular, on the LasHeR dataset, our method achieves 2.3\% and 0.9\% improvements in PR and SR, respectively, compared to FlexTrack. However, we note that our tracker performs slightly worse than LRPD, a method specifically designed for RGB-T tracking. This is likely due to LRPD utilizing the more powerful DINOv2~\cite{oquab2023dinov2} feature extractor. Nevertheless, while LRPD is limited to RGB-T tracking tasks, our method can be applied to various tracking tasks such as RGB-E and RGB-D under a unified framework and parameter settings, demonstrating superior versatility and scalability.

\begin{table}[htbp]\small
  \centering
\renewcommand\arraystretch{1.2}  
     \caption{Overall performance on COESOT. Results are reported in percentage (\%). The $*$ indicates that our method performs full fine-tuning.}
     \setlength{\tabcolsep}{4.5mm}{ \begin{tabular}{l|c|cc}
    \toprule
  \multicolumn{1}{c|}{\multirow{2}[2]{*}{Method}} & \multicolumn{1}{c}{\multirow{2}[2]{*}{Source}} & \multicolumn{2}{c}{COESOT} \\
   &  & \multicolumn{1}{c}{PR} & \multicolumn{1}{c}{AUC} \\
    \midrule
  SDMoEA*-L & \textbf{Ours} &  \textbf{85.1}& \textbf{70.3}  \\
  SDMoEA*-B & \textbf{Ours} &84.4  & 69.3\\ 
  SDMoEA-L &  \textbf{Ours}& 83.2 & 68.8 \\
  SDMoEA-B & \textbf{Ours} &83.6 &68.4\\
  DS-MESA~\cite{shao2024dynamic}    & PRCV'24    & 77.5  &69.1 \\ 
  EventTPT~\cite{xia2025towards}    &  ICRA'25  & 73.0  & 64.7\\ 
  CMDTrack~\cite{cmdtrack}     & PAMI'25     &74.8   &65.7  \\
  CEUTrack~\cite{tang2022revisiting}    &  PR'25    & 76.0  &62.7\\
  CRSOT~\cite{zhu2025crsot}    &  TMM'25   & 75.1  &60.8 \\
  OSTrack-FE~\cite{ye2022joint}    & ECCV'22     & 70.7  &59.0 \\   
  ToMP101-FE~\cite{mayer2022transforming}    & CVPR'22      & 71.6 &59.9 \\ 
\bottomrule
    \end{tabular}}
  \label{tab:coesot}%
\end{table}%

\begin{table*}[htbp]\small
  \centering
\renewcommand\arraystretch{1.2}  
 \caption{Overall performance on VOT-RGBD2022 and DepthTrack. Results are reported in percentage (\%). PEFT and FFT represent parameter-efficient fine-tuning and full fine-tuning, respectively. The $*$ indicates that our method performs full fine-tuning.}
     \setlength{\tabcolsep}{2.5mm}{ \begin{tabular}{c|c|l|c|ccc|ccc}
    \toprule
    \multicolumn{2}{c|}{\multirow{2}[2]{*}{Method Type}} & \multicolumn{1}{c|}{\multirow{2}[2]{*}{Method}} & \multicolumn{1}{c|}{\multirow{2}[2]{*}{Source}} & \multicolumn{3}{c|}{VOT-RGBD2022} & \multicolumn{3}{c}{DepthTrack} \\
   \cmidrule{5-10} \multicolumn{2}{c|}{} &      &      & \multicolumn{1}{c}{EAO} & \multicolumn{1}{c}{Acc.} &\multicolumn{1}{c|}{Rob.}& \multicolumn{1}{c}{F-score} & \multicolumn{1}{c}{Re} & \multicolumn{1}{c}{Pr}\\
    \midrule
    \multicolumn{1}{c|}{\multirow{19}[4]{*}{\tabincell{c}{Multi-modal \\ Trackers}}} & \multicolumn{1}{c|}{\multirow{11}[2]{*}{PEFT}} & SDMoEA-L &  \textbf{Ours} & \textbf{78.4} & \textbf{83.0} & 94.2 &\textbf{69.1} &\textbf{68.4}&\textbf{68.8} \\
         &      & SDMoEA-B & \textbf{Ours} & 78.3 & 81.9&\textbf{94.9} &64.7  &65.2&65.0 \\
         &      & XTrack-L~\cite{tan2024xtrack}    &  ICCV'25    &74.0    &82.8 & 88.9 &64.8 &64.3&65.4\\
         &      & XTrack-B~\cite{tan2024xtrack}    &  ICCV'25 &74.0      & 82.1   &  88.8   & 61.5&62.0&61.8\\         
         &      & SeqTrackv2-L~\cite{chen2023unified}    &  arXiv'23    &74.8  &82.6 & 91.0  & 62.3&62.6&62.5 \\ 
         &      & SeqTrackv2-B~\cite{chen2023unified}   &  arXiv'23    &75.5 & 81.9 &91.8 &59.8 &60.0 &59.6\\                
         &      & OneTracker~\cite{hong2024onetracker} & CVPR'25     &72.7&81.9 &87.2 &60.9 &60.4&60.7\\
         &      & EMTrack~\cite{10747517}   &  TCSVT'25   &69.7   & 80.6   &84.4 &58.3 &58.0&58.5\\         
         &      &Un-Track~\cite{wu2024single}    &  CVPR'24    &72.1    & 82.0  & 86.9   &61.0 &60.8 &61.1\\
         &      &SDSTrack~\cite{hou2024sdstrack}    &  CVPR'24    &72.8&81.2  &88.3  &61.4&60.9&61.9\\       
         &      &ViPT~\cite{zhu2023visual}    &  CVPR'23    &72.1 & 81.5  &87.1&59.4&59.6&59.2\\
\cmidrule{2-10}         & \multirow{9}[2]{*}{FFT} & SDMoEA*-L & \textbf{Ours} &77.5 & 81.6 &94.5 & \textbf{67.6}&\textbf{67.4}&\textbf{67.5}\\
         &      & SDMoEA*-B & \textbf{Ours} & \textbf{78.7} &82.2 & \textbf{95.0} &65.9 &65.7&65.8\\
         &      & FlexTrack~\cite{tan2025you} &   ICCV'25  & 78.0 & 83.8 &93.1 & 67.0 &66.9&67.1\\         
         &      & SUTrack-L~\cite{chen2025sutrack}  & AAAI'25& 76.6 & 83.5& 92.2 & 66.4 & 66.4& 66.5\\
         &      & SUTrack-B~\cite{chen2025sutrack} & AAAI'25 &76.6 & \textbf{83.9} & 91.4& 64.4 & 64.2&64.6\\
         &      & STTrack~\cite{hu2025exploiting} &    AAAI'25     &77.6 & 82.5 & 93.7 &63.3 &63.4 &63.2\\
         &      &  APTrack~\cite{aptrack}    & TAI'25    &77.4    &82.1  &93.4  & 62.1 &61.9&62.3\\
         &      & CMDTrack~\cite{cmdtrack}     & PAMI'25     & -     & -   & -     & 59.1 &60.7 &59.8\\
         
    \midrule
    \multirow{7}[3]{*}{\tabincell{c}{RGBD \\ Trackers}} & \multicolumn{1}{c|}{\multirow{1}[2]{*}{PEFT}}    & MixFormer\_3DPT~\cite{Li_2024_ACCV}     & ACCV'24     &70.0     & 81.8   &84.7     & 62.0 &61.0&62.9\\

\cmidrule{2-10}         & \multirow{5}[1]{*}{FFT}      & TABBTrack~\cite{ying2025temporal}     &PR'24   & 72.2 &82.1	&87.4		&61.8	&61.5 &62.2			\\
         &      & AMATrack~\cite{10623547}     &TIM'24   &- &-	&-		&61.8	&60.7 &62.9  \\
         &      & DepthRefiner~\cite{lai2024depthrefiner}     & ICME'24     &60.3 & 79.7	&73.3	&51.0 &50.7 &51.3		\\
         &      & RDT-TEF~\cite{gao2025rgb}     &KBS'25   & 71.0 &80.7	&87.7		&57.4	&53.8 &61.5 \\    
         &      & SSLTrack~\cite{zhu2024self}     &PR'24   &- &-	&-		&52.5	&49.1 &56.5  \\             
\bottomrule
    \end{tabular}}
  \label{tab:rgbd}%
\end{table*}%
\subsubsection{Evaluation on RGB-E Tracking Datasets}
As shown in Table~\ref{tab:rgbe}, we conduct a comprehensive performance comparison on the VisEvent dataset. compared with various parameter-efficient fine-tuning (PEFT) multi-modal tracking methods, our method achieves the best results on the PR metric. Notably, even compared with all types of tracking methods, our FFT variant still performed best on the PR metric, while on the AUC metric it was only 0.3\% lower than the most competitive method. Furthermore, to provide a more comprehensive comparison, we also conduct experimental comparisons on the large-scale COESOT dataset. As shown in Table~\ref{tab:coesot}, our method achieves the best tracking performance on both PR and AUC metrics. The above experimental results fully validate the effectiveness and superiority of the proposed method.



\subsubsection{Evaluation on RGB-D Tracking Datasets}
To validate the effectiveness of the proposed method on RGBD datasets, we evaluate our approach alongside existing methods on two widely used RGBD tracking datasets: VOT-RGBD2022~\cite{kristan2022tenth} and DepthTrack. As shown in the Table~\ref{tab:rgbd}, on the DepthTrack dataset, both versions of our method achieve the best performance across all three metrics: F-score, Precision, and Recall. Specifically, our PEFT version outperformed all comparable methods, achieving 2.1\%, 1.5\%, and 1.7\% higher F-score, Precision, and Recall, respectively, compared to the second-best FlexTrack method. On the VOT-RGBD2022 dataset, our method also demonstrates outstanding performance, particularly achieving the best results in both EAO and Rob. metrics. 
As summarized in Table~\ref{tab:rgbt},Table~\ref{tab:rgbe}, Table~\ref{tab:coesot}, and Table~\ref{tab:rgbd}, our method achieves highly competitive performance across RGBT, RGBE, and RGBD datasets, demonstrating robust and generalizable capabilities.

\begin{table*}[htbp]
  \centering  
  \caption{Ablation studies of different components of the proposed method on the testing sets of LasHeR and VisEvent. Results are reported in percentage (\%).}
\belowrulesep=0pt
\aboverulesep=0pt
  \renewcommand\arraystretch{1.9}
 \setlength{\tabcolsep}{2.5mm}{  
  \begin{tabular}{c | c c c c|c c| c c|c c}
    \hline
    \multirow{1}[5]{*}{Method}& \multirow{1}[5]{*}{SDMoE} & \multirow{1}[5]{*}{MFF} & \multirow{1}[5]{*}{\tabincell{c}{GSAHF-MHG}} & \multirow{1}[5]{*}{\tabincell{c}{GSAHF-Gram}} & \multirow{1}[5]{*}{Params}& \multirow{1}[5]{*}{GFLOPs}
    & \multicolumn{2}{c|}{LasHeR}     & \multicolumn{2}{c}{VisEvent} \\ 
    \cmidrule{8-11}      &&  & &&& & {PR}& {SR} &{PR} &{SR}   \\   
    
    \hline
    \textrm{I}    & & &  & &97.43 M&170.17& 60.9&47.9&66.9&45.0\\
    \midrule   
    
    \textrm{II}   &\Checkmark &  & && 105.88 M&173.15 &74.3  &59.0  &79.3&60.4 \\
    \textrm{III}   & &\Checkmark  & && 100.97 M&174.27 &65.6  & 51.3 &77.4 &57.6 \\   
    \textrm{IV}   & &  &\Checkmark && 102.15 M& 175.63& 68.0 & 53.8 &78.5 &59.0 \\ 
    \textrm{V}   & &  & & \Checkmark& 101.56 M&176.99 &70.8  & 56.1& 78.9&59.1 \\ 
    \midrule
    
    \textrm{VI}  &\Checkmark&\Checkmark  & & &109.42 M& 177.23 &75.3 &59.6&79.6&60.7\\
    \textrm{VII}  &\Checkmark&  &\Checkmark & &110.60 M&178.59 &75.1  & 59.2 &79.8 &60.5 \\       
    \textrm{VIII}  &\Checkmark&  & &\Checkmark &110.02 M& 179.95&74.5 &58.8 &80.0 & 60.4\\   \midrule 
    \textrm{IX}   &\Checkmark  &\Checkmark  &\Checkmark &  &114.15 M& 182.67&75.4 &59.8 &80.2&60.8\\       
    \textrm{X}   &\Checkmark &\Checkmark & &\Checkmark &113.56 M& 184.03&75.0&59.2 &80.3&61.0\\
    \textrm{XI}   &\Checkmark & &\Checkmark &\Checkmark &114.74 M& 185.39&75.9&60.1 &79.8&60.9\\    
    \midrule
    \textrm{XII}  &\Checkmark &\Checkmark &\Checkmark &\Checkmark &118.28 M& 189.47&76.5&60.3&80.3&61.2\\   %
    
   
    \hline
  \end{tabular}}
\label{tb::AS}
\end{table*}

\begin{table}[htbp]\small
  \centering
\renewcommand\arraystretch{1.6}  
     \caption{Performance comparison of embedding SDMOE in odd and even layers of the backbone network on LasHeR and VisEvent. Results are reported in percentage (\%). }
     \setlength{\tabcolsep}{2.1mm}{ \begin{tabular}{l|ccc|ccc}
    \toprule
  \multicolumn{1}{c|}{\multirow{2}[2]{*}{Method}}  & \multicolumn{3}{c|}{LasHeR} & \multicolumn{3}{c}{VisEvent}\\
  \cmidrule{2-7}    & \multicolumn{1}{c}{PR} & \multicolumn{1}{c}{NPR} & \multicolumn{1}{c|}{SR} & \multicolumn{1}{c}{PR} & \multicolumn{1}{c}{NPR} & \multicolumn{1}{c}{SR}\\
    \midrule
  SDMoEA-odd  &  75.3&71.3& 59.3 &80.1&75.7&60.9 \\
  SDMoEA-even  &71.7  &68.2 & 56.7 &80.4&75.1&60.8\\ 
  SDMoEA  & 76.5 &72.6 & 60.3 &80.3&75.5 &61.2\\
\bottomrule
    \end{tabular}}
  \label{tab:Sdmoe-oddeven}%
\end{table}%

\begin{table}[htbp]\small
  \centering
\renewcommand\arraystretch{1.6}  
     \caption{Performance comparison of sparse MoE and dense-shared of SDMoE on LasHeR and VisEvent. Results are reported in percentage (\%).}
     \setlength{\tabcolsep}{2.1mm}{ \begin{tabular}{l|ccc|ccc}
    \toprule
  \multicolumn{1}{c|}{\multirow{2}[2]{*}{Method}}  & \multicolumn{3}{c|}{LasHeR} & \multicolumn{3}{c}{VisEvent}\\
  \cmidrule{2-7}    & \multicolumn{1}{c}{PR} & \multicolumn{1}{c}{NPR} & \multicolumn{1}{c|}{SR} & \multicolumn{1}{c}{PR} & \multicolumn{1}{c}{NPR} & \multicolumn{1}{c}{SR}\\
    \midrule
  SDMoEA-spare  &  74.0&70.4& 58.5 &79.2&74.9&60.4 \\
  SDMoEA-dense  &74.9 &71.5 & 59.1 &80.5&75.9&61.2\\ 
  SDMoEA  & 76.5 &72.6 & 60.3 &80.3&75.5 &61.2\\
\bottomrule
    \end{tabular}}
  \label{tab:Sdmoe-sd}%
\end{table}%

\subsubsection{Ablation Study}
To verify the effectiveness of each component of the proposed method, we conduct several ablation experiments. We use the same pre-training model and hyperparameter configurations in all experiments with the same number of rounds of training for ensuring the fairness of the experiments. The baseline approach first extends the ODTrack to dual-modal input, extracting features from different modalities through two independent branches. These features are then concatenated at the final layer of the feature extractor to construct a multi-modal tracker. Subsequently, the model parameters of the ODTrack are directly loaded and tested on multi-modal data. 

\textbf{Analysis of SDMoEA Core Components.} As shown in Table~\ref{tb::AS}, the SDMoE, MFF, GSAHF-MHG and GSAHF-Gram denote the sparse-dense mixture of experts module, multi-level feature fusion module, the hypergraph fusion part of gram-based semantic alignment hypergraph fusion module and the gram-based semantic alignment part of gram-based semantic alignment hypergraph fusion module, respectively.

As shown in Table~\ref{tb::AS} (I,II, III, IV, V), SDMoE, MFF, GSAHF-MHG, and GSAHF-Gram have a significant improvement compared to baseline, which is a strong demonstration of the effectiveness of the four modules. As shown in Table.\ref{tb::AS} (II, VI, VII, VIII), compared to using SDMoE alone, combining it with MFF, GSAHF-MHG, or GSAHF-Gram respectively yields varying degrees of performance improvement across two multi-modal tracking datasets.
As shown in Table~\ref{tb::AS}(VI, IX, X), on the LasHeR dataset, SDMoE+MFF+GSAHF-MHG (Method IX) yields only marginal performance gains over the SDMoE+MFF (Method VI), while SDMoE+MFF+GSAHF-Gram (Method X) even results in a slight deterioration in performance. We attribute this to the inherent differences between multi-level, multi-modal features, which hinder effective feature modeling with SDMoE+MFF+GSAHF-MHG or SDMoE+MFF+GSAHF-Gram. Further analysis of Tables~\ref{tb::AS}(VII, VIII, XI) reveals that combining GSAHF-Gram with GSAHF-MHG within the SDMoE framework yields significant performance gains. This demonstrates that first aligning multi-level multi-modal features using GSAHF-Gram, followed by high-order modeling of the aligned features via GSAHF-MHG, substantially enhances feature representational capacity.
Finally, as shown in Table~\ref{tb::AS}(XII), integrating all modules achieves optimal tracking performance.

\textbf{Analysis of Parameter and Computational Cost.}
Additionally, we present the parameter quantities and computational costs for variant models across different components. As shown in Table~\ref{tb::AS}(II), our core module SDMoE delivers the most significant performance gains compared to baseline methods, requiring only an additional 8.45 million parameters and 2.98 GFLOPs of computational effort. Owing to the structural design advantages of SDMoE, its computational load increases by merely 1.75\% despite an 8.67\% rise in parameters. Upon integrating all modules, the final model exhibits an increase of 20.28 million parameters and 19.3 GFLOPs in computational load.

\textbf{Analysis of SDMoE Number.} To evaluate the impact of the number of proposed SDMoE modules on model performance, we insert the module into both odd and even layers of the backbone network for comparison. As shown in Table~\ref{tab:Sdmoe-oddeven}, on the LasHeR dataset, inserting the SDMoE module into either odd or even layers of the backbone network yield inferior results compared to full-layer insertion, with a noticeable performance gap. On the VisEvent dataset, the full-layer insertion approach achieves optimal performance on the SR metric. Although it slightly underperformed on the PR and NPR metrics compared to other configurations, considering the overall performance across both datasets, inserting the SDMoE module across all layers yields more balanced and superior comprehensive results.
\begin{figure*}[b]
\centering 
\includegraphics[width=0.95\textwidth]{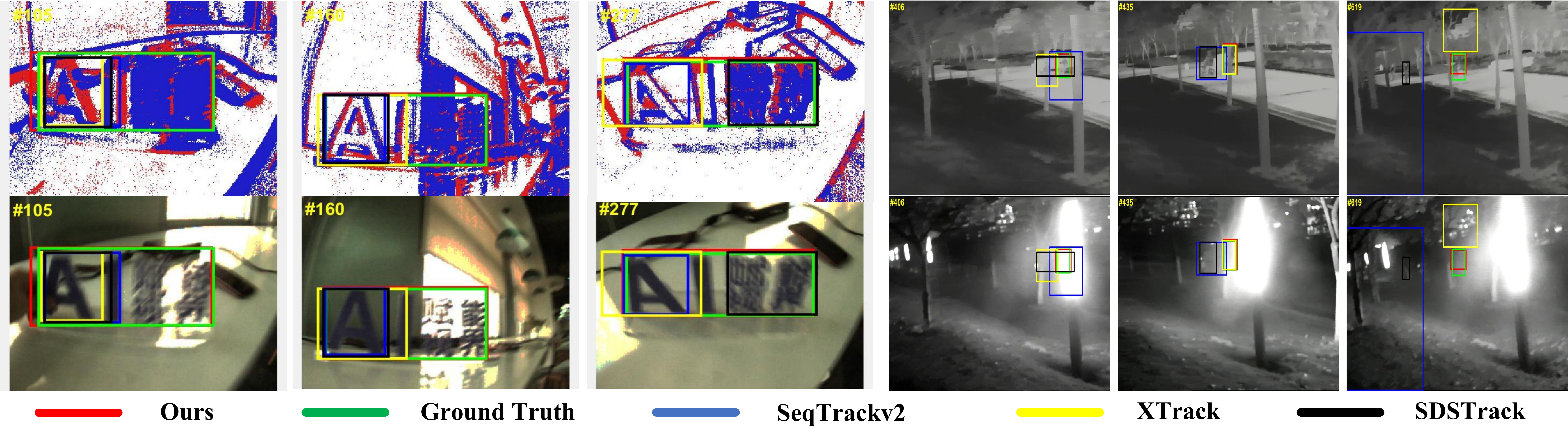}
   \caption{Visualization results on multi-modal tracking datasets.}
   \label{fig:qc}
\end{figure*}

\textbf{Analysis of Sparse MoE and Dense-shared MoE in SDMoE.}
To evaluate the impact of sparse MoE versus dense-shared MoE on multi-modal tracking performance within the SDMoE module, we conduct two comparative experiments. Here, SDMoEA-sparse denotes the use of sparse MoE alone during the embedding stage, while SDMoEA-dense denotes the use of dense-shared MoE alone. The experimental results are presented in Table~\ref{tab:Sdmoe-sd}, demonstrating that SDMoEA-dense significantly outperforms SDMoEA-sparse on both the LasHeR and VisEvent datasets. This indicates that mining shared features across modalities is more crucial for improving tracking performance.
Furthermore, on the LasHeR dataset, the architecture combining sparse MoE with dense-shared MoE demonstrate markedly superior performance compared to single-component models. Although SDMoEA-dense achieves slightly better PR and NPR metrics than the combined model (i.e. SDMoEA) on the VisEvent dataset, the combined model exhibited balanced overall performance across both datasets.
\begin{figure*}[htbp]
\centering 
\includegraphics[width=0.95\textwidth]{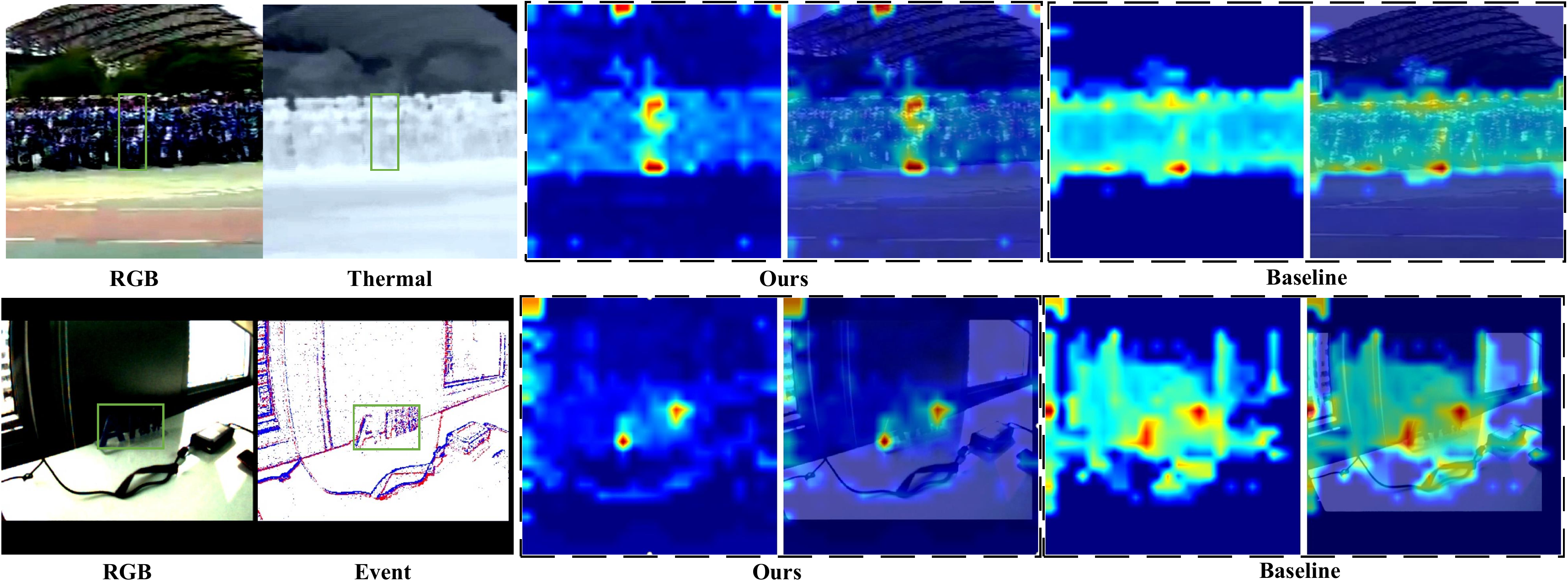}
\caption{Visualization of feature maps. The first to sixth columns are the RGB input, Thermal or Event input, the feature map of our method, class activation mappingand of our method, the feature map of baseline method, class activation mapping of baseline method, respectively.}
   \label{fig:feature_map}
\end{figure*}

\subsection{Qualitative Comparisons}
To intuitively compare our tracking performance with that of other trackers, we present a qualitative comparison between our algorithm and three trackers, on two representative multi-modal video sequences, as shown in Fig.~\ref{fig:qc}.
These video sequences encompass both RGB-E and RGB-T data types, which include a variety of typical challenging scenarios such as fast motion, similar object, low illumination, strong glare, occlusion, deformation and background clutter. 
As shown in Fig. \ref{fig:qc}, our method is more robust than other algorithms under complex challenge scenarios. For example, in the RGB-E sequence (the left half of Fig.~\ref{fig:qc}), it has fast motion, background clutter and deformation challenges, our algorithm can still run well compared with other algorithms. For another example, in the RGB-T sequence (the right half of Fig.~\ref{fig:qc}), it has similar object, low illumination, strong glare, occlusion and background clutter. These challenges cause significant degradation of the target's appearance characteristics and render it difficult to capture effectively in the visible modality. In such a difficult challenge scenario, our algorithm can still track the target well, which shows that our algorithm can really mine the complementary features between modalities for more robust multi-modal tracking.
In addition, we perform a visual analysis of the feature maps for the search region. As illustrated in Fig.~\ref{fig:feature_map}, under both RGB-E and RGB-T modalities, the feature maps generated by baseline methods exhibit significant limitations, being susceptible to interference from factors such as similar object or low illumination. In contrast, the proposed method demonstrates enhanced robustness and generalisation capability, yielding clearer and sharper feature maps that effectively suppress interference from modally similar object and low illumination environments.

\section{Conclusion}
In this paper, we address the challenges of parameter-efficient fine-tuning for unified multi-modal visual tracking. The inherent heterogeneity across modalities makes it difficult for existing methods to effectively represent both modality-specific and shared information within a unified network framework and parameter set. To overcome these limitations, we propose a novel Sparse-Dense Mixture of Experts Adapter (SDMoEA) framework.
Our core innovation is the SDMoE module, which combines a sparse MoE for capturing modality-specific features and a novel serial-parallel dense-shared MoE for modeling cross-modal shared information. This design significantly enhances representational capacity and efficiency within a unified framework and parameter set. Furthermore, we design the Gram-based Semantic Alignment Hypergraph Fusion (GSAHF) module to better model the high-order correlations in multi-level, multi-modal feature fusion, leading to more robust feature representations. Extensive experiments on major multi-modal tracking benchmarks, including LasHeR, RGBT234, VTUAV, VisEvent, COESOT, DepthTrack, and VOT-RGBD2022, demonstrate that our method achieves highly competitive performance compared to other state-of-the-art parameter-efficient fine-tuning techniques. These results provide strong validation of the efficacy of our approach.
Future work will explore the integration of more diverse modalities and the extension of the SDMoEA framework to other multi-modal vision tasks beyond tracking. The proposed efficient shared and specific feature modeling approach presents a promising direction for generalised parameter-efficient multi-modal learning.

\bibliographystyle{IEEEtran}
\bibliography{main}
\end{document}